\def\year{2020}\relax
\documentclass[letterpaper]{article} 
\usepackage{aaai20}  
\usepackage{times}  
\usepackage{helvet} 
\usepackage{courier}  
\usepackage[hyphens]{url}  
\usepackage{graphicx} 
\usepackage{graphicx}  
\usepackage{amsmath}
\usepackage{amssymb}
\usepackage{dsfont}
\usepackage{booktabs}
\usepackage{pifont}
\usepackage{multirow}
\usepackage[ruled]{algorithm2e}
\SetKwComment{Comment}{$\triangleright$\ }{}
\usepackage{subcaption}

\DeclareMathOperator*{\argmax}{arg\,max}

\usepackage{xcolor}

\title{CD-UAP: Class Discriminative Universal Adversarial Perturbation}
\author{Chaoning Zhang\textsuperscript{\rm *}, Philipp Benz\textsuperscript{\rm *}, Tooba Imtiaz, In-So Kweon \\
Korea Advanced Institute of Science and Technology (KAIST), South Korea \\ 
 \textsuperscript{\rm *} Equal contribution \\
\tt\small{chaoningzhang1990@gmail.com}, \tt\small\{pbenz, timtiaz, iskweon77\}@kaist.ac.kr}
\begin{document}

\maketitle

\begin{abstract}
A single universal adversarial perturbation (UAP) can be added to all natural images to change most of their predicted class labels. It is of high practical relevance for an attacker to have flexible control over the targeted classes to be attacked, however, the existing UAP method attacks samples from all classes. In this work, we propose a new universal attack method to generate a single perturbation that fools a target network to misclassify only a chosen group of classes, while having limited influence on the remaining classes. 
Since the proposed attack generates a universal adversarial perturbation that is discriminative to targeted and non-targeted classes, we term it class discriminative universal adversarial perturbation (CD-UAP).
We propose one simple yet effective algorithm framework, under which we design and compare various loss function configurations tailored for the class discriminative universal attack. 
The proposed approach has been evaluated with extensive experiments on various benchmark datasets. Additionally, our proposed approach achieves state-of-the-art performance for the original task of UAP attacking all classes, which demonstrates the effectiveness of our approach.
\end{abstract}

\section{Introduction}
Deep neural networks (DNNs) are known to be vulnerable to malicious attacks of visually inconspicuous adversarial examples~\cite{szegedy2013intriguing,qiu2019review}. The reason behind this intriguing DNN property is not fully understood~\cite{goodfellow2014explaining,tanay2016boundary}, however, researchers have exploited this phenomenon to come up with various attack methods~\cite{akhtar2018threat}.

\begin{figure}
\includegraphics[width=0.47\textwidth]{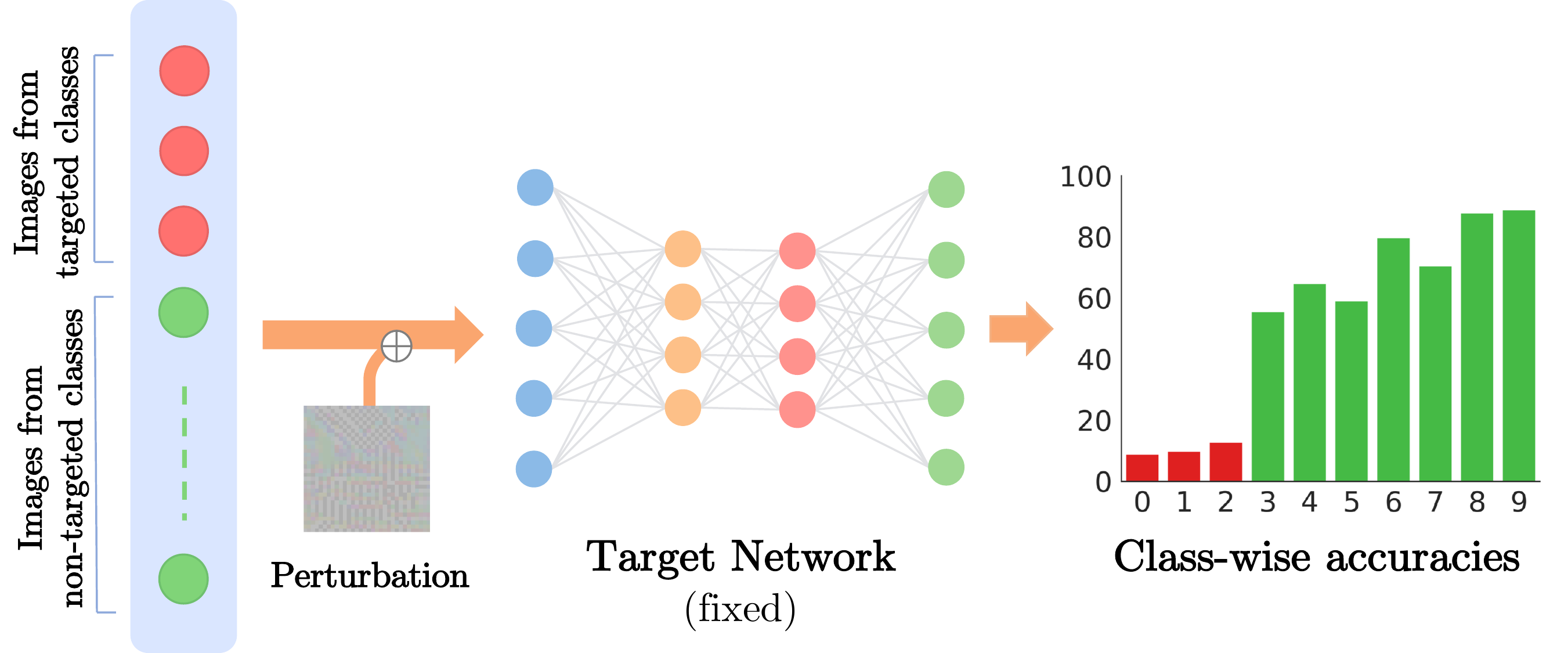}
\caption{Class Discriminative Universal Adversarial Perturbation (CD-UAP). After adding the perturbation, the model performance on a subset of classes (targeted classes) is significantly reduced, while the influence on the non-targeted classes is limited. We demonstrate this with the results achieved on CIFAR10 dataset.}
\label{fig:teaser}
\end{figure}

The existing adversarial attack methods can be categorized into image-dependent attacks and image-agnostic attacks~\cite{akhtar2018threat}. Image-dependent attacks craft perturbations that can fool the network for one specific input image. Due to the image-dependent nature, the perturbations have to be crafted individually for each target image~\cite{szegedy2013intriguing}. On the other hand, image-agnostic attacks, also called universal attacks, craft one single perturbation for converting every image from a data distribution into an adversarial example~\cite{moosavi2017universal}. The universal nature has the practical benefit that the perturbations can be crafted in advance, which makes them more convenient to use for an attacker. However, the existing universal attacks~\cite{moosavi2017universal,neekhara2019universal} fool the network for samples from every class, which can lead to obvious network misbehavior and raise suspicion. Consequently, it can be of practical relevance for an attacker to have control over the classes to attack. A natural question arises whether it is possible to craft a universal perturbation that fools the network only for certain classes while having minimal influence on other classes.

In this work, we propose the task of class discriminative UAP (CD-UAP) as shown in Figure~\ref{fig:teaser}. To distinguish our approach from the original UAP by~\citeauthor{moosavi2017universal} we term their task of a UAP attacking all classes All-Classes UAP (AC-UAP). AC-UAP can be seen as a special case of CD-UAP when all classes are targeted. Nonetheless, in this work by default, CD-UAP does not attack all classes. Ideally, the proposed CD-UAP negatively affects only the targeted classes. We argue that this property makes the CD-UAP more covert than the AC-UAP. 
Strictly speaking, the proposed attack falls no longer under the category of universal attacks, since it does not fool a network for samples from every class. We still term it universal attack, since the perturbation is still applied to all image samples~\cite{moosavi2017universal}, while aiming to misclassify only the targeted classes. 

The overall objective of the proposed CD-UAP can be decomposed into two parts: maximizing the attack success rate for the targeted classes, while minimizing the influence of the perturbation on the non-targeted classes. In practice, these two goals contradict each other, and an inevitable trade-off emerges, therefore, it is a non-trivial task to craft CD-UAPs. A na\"{i}ve approach is to apply the existing UAP methods to only the targeted classes. However, perturbations crafted only on the targeted classes also successfully fool the network for samples from the non-targeted classes, implying that na\"{i}vely targeting a subset of classes by UAP~\cite{moosavi2017universal} cannot achieve the desired attack behavior. Moreover, since the perturbations are noise by nature, it is theoretically impossible for them to have no influence on images of the non-targeted classes. Nonetheless, it is possible to limit such influence. Recognizing the trade-off between the two contradicting goals, we propose a simple yet effective algorithm framework that explicitly addresses the targeted and non-targeted classes with separated loss functions. Under this framework, we design and compare various loss function variants to explore the optimal combination for this task. The proposed approach has been evaluated on various benchmark datasets for different DNNs.
To sum up, our contributions are as follows:
\begin{itemize}
\item We first show the existence of a class-discriminative universal adversarial perturbation (CD-UAP), that allows flexible control over the targeted classes to attack, on several benchmark datasets: CIFAR10, CIFAR100 and ImageNet.
\item Identifying the limitations of the standard UAP attack method, we propose an efficient algorithm framework, explicitly handling images from targeted classes and non-targeted classes with separate loss functions for promoting class discrimination.
\item Under the proposed framework, we carefully design and compare various loss function configurations while specifically taking into account the balance between the two contradicting goals.
\item Our approach achieves state-of-the-art performance for the task of AC-UAP, which demonstrates the effectiveness of our approach.
\end{itemize}

\section{Related Work}
\citeauthor{szegedy2013intriguing} first reported the intriguing property of DNN vulnerability to maliciously crafted small perturbations~\cite{szegedy2013intriguing}. Since then, adversarial attacks and defenses have become an active research field. The readers can refer to~\cite{qiu2019review,yuan2019adversarial} for a comprehensive review and we summarize only the works related to adversarial attacks~\cite{akhtar2018threat} in this section. There are different ways to categorize attacks, such as targeted and non-targeted attacks, or white-box and black-box attacks. Here we categorize them into image-dependent attacks and image-agnostic attacks.

\subsection{Image-Dependent Adversarial Perturbations}
\citeauthor{szegedy2013intriguing} proposed to use box-constrained L-BFGS to generate perturbations that can fool a network~\cite{szegedy2013intriguing}. The Fast Gradient Sign Method (FGSM)~\cite{goodfellow2014explaining}, which is a one-step attack, was then proposed to update the perturbations via the direction of the gradients. Iterative FGSM (I-FGSM)~\cite{kurakin2016adversarial}, iteratively performs the FGSM attack. In each iteration, only a fraction of the allowed noise limit is added, which contributes to its higher attack effect compared to FGSM. A momentum term, which was previously used to train DNNs, is introduced in Momentum I-FGSM to obtain smoother gradient update directions~\cite{dong2018boosting}. To improve transferability, Variance-Reduced I-FGSM~\cite{wu2018understanding} utilizes the averaged gradient of images with Gaussian noise which replaces the gradient of the original image. DeepFool~\cite{moosavi2016deepfool} crafts perturbations iteratively by updating the gradient with respect to the model’s decision boundaries. Other widely used powerful attacks include the Carlini and Wagner (C\&W) attack~\cite{carlini2017towards}, and projected gradient descent (PGD)~\cite{madry2017towards}, which have been empirically shown to be strong attacks.
Image-dependent attacks target a single image and their main limitation therefore is that they cannot be computed in advance, but instead have to be computed on the spot. 

\subsection{Image-Agnostic Adversarial Perturbations}
Image-agnostic adversarial perturbations, also widely known as universal adversarial perturbations (UAP), were first proposed to construct one single perturbation which is able to attack most images from a certain data distribution~\cite{moosavi2017universal}. 
\citeauthor{khrulkov2018art} proposed to craft a UAP based on the Jacobian matrices of the networks hidden layers, resulting in interesting visual patterns~\cite{khrulkov2018art}. A data-free UAP was proposed to maximize the feature change caused by the perturbation~\cite{Mopuri2017datafree,mopuri2018generalizable}. UAPs were also extended beyond classification to the field of semantic segmentation~\cite{metzen2017universal}. In addition, there have also been attempts to craft UAPs using generative models~\cite{poursaeed2018generative}, as well as in in real-world scenarios~\cite{brown2017adversarial,athalye2017synthesizing,sharif2017adversarial}. UAPs have the advantage that they can be computed in advance, which can be more practical for a potential attacker. However, existing universal attacks cannot give an attacker the freedom of control over the targeted classes. In this work, we identify this limitation and propose class discriminative universal adversarial perturbations (CD-UAP).

\section{Class Discriminative Universal Attack}
\subsection{Problem Formulation}
Conceptually, we aim to craft a single perturbation which only attacks samples from a group of targeted classes, while limiting the perturbation influence on samples from the other classes. This objective involves two contradicting goals: maximizing the attack rate on samples from the targeted classes and minimizing the accuracy drop for the non-targeted classes. In this section, we first restate the formulation of AC-UAP and derive a formulation for the proposed CD-UAP.

Let $X \in \mathds{R}^d$ be a data distribution and $\hat{F}$ be a classifier, which maps input images $x \sim X$ to an estimated label $\hat{F}(x)$. Universal perturbations seek a perturbation vector $\delta \in \mathds{R}^d$ that fools the classifier $\hat{F}$ on most data points~\cite{moosavi2017universal}, which can be illustrated as
\begin{equation*}
    \hat{F}(x+\delta) \neq \hat{F}(x)\text{ for \textit{most} }x \sim X.
\end{equation*}
The perturbations are constrained to be smaller than a certain magnitude $\epsilon$  
\begin{equation*}
    ||\delta ||_p \leq \epsilon 
\end{equation*}
to be visually imperceptible to humans.
The existing AC-UAP technique ideally aims to fool the model for all image samples. We argue that the behavior of a network under such an attack is suspicious and can easily catch attention of a user. In order to design a more \textit{stealthy} attack, we propose the class discriminative universal attack, through which an attacker can choose a set of targeted classes $S$. The algorithm then searches for a perturbation vector $\delta$ which fools images belonging to $S$ ($x_{\text{t}} \sim X_{\text{t}}$), while limiting its influence on the images belonging to the non-targeted classes ($x_{\text{nt}} \sim X_{\text{nt}}$). 
Therefore, the formulation of AC-UAP can be extended to fit the objective of CD-UAP as follows:
\begin{align*}
    \hat{F}(x_{\text{t}} + \delta) &\neq \hat{F}(x_{\text{t}}) \text{ for \textit{most} } x_{\text{t}} \sim X_{\text{t}} \\
    \hat{F}(x_{\text{nt}} + \delta) &= \hat{F}(x_{\text{nt}}) \text{ for \textit{most} } x_{\text{nt}} \sim X_{\text{nt}},
\end{align*}
while keeping the perturbation magnitude limited to a certain threshold $\epsilon$, i.e. $||\delta ||_p \leq \epsilon$.

As a reference value, we report the initial classification accuracies for the targeted classes $Acc_{\text{t}}$ and that for the non-targeted classes $Acc_{\text{nt}}$.
\begin{align*}
    Acc_{\text{t}} &= Acc(\hat{F}(x_{\text{t}}),y) \text{ for } x_{\text{t}} \sim X_{\text{t}} \\
    Acc_{\text{nt}} &= Acc(\hat{F}(x_{\text{nt}}),y) \text{ for } x_{\text{nt}} \sim X_{\text{nt}},
\end{align*}
where $y$ indicates the ground truth label.
Furthermore, in our experiments we use the absolute accuracy drop ($AAD$) as an evaluation metric, which is defined for the targeted classes and the non-targeted classes as:
\begin{align*}
    AAD_{\text{t}} &= Acc_{\text{t}}(\hat{F}(x_{\text{t}}), y) - Acc_{\text{t}}(\hat{F}(x_{\text{t}} + \delta), y) \\
    AAD_{\text{nt}} &= Acc_{\text{nt}}(\hat{F}(x_{\text{nt}}), y) - Acc_{\text{nt}}(\hat{F}(x_{\text{nt}} + \delta), y),
\end{align*}
where, according to the defined objective, higher $AAD_{\text{t}}$ and lower $AAD_{\text{nt}}$ are desired. The two metrics can be combined into one overall metric, i.e., the absolute accuracy drop gap $\Delta_{AAD}$ as:
\begin{equation*}
    \Delta_{AAD} = AAD_{\text{t}} - AAD_{\text{nt}}.
\end{equation*}

\subsection{Algorithm Framework}
\begin{algorithm}[t]
    \small
    \SetAlgoLined
    \DontPrintSemicolon
    \SetKwInput{KwInput}{Input}
    \SetKwInput{KwOutput}{Output}
    \SetKwFunction{FAdam}{ADAM}
    \KwInput{Data distribution $X$, Classifier $\hat{F}$, Loss function $\mathcal{L_{\text{t}}}$ and $\mathcal{L_{\text{nt}}}$, Batch size $b$, Number of iterations $N$, hyper-parameters $\alpha$ and $\beta$}
    \KwOutput{Class discriminative universal perturbation vector $\delta$}
    $X_{\text{t}}, X_{\text{nt}} \subseteq X$ \\
    $\delta \leftarrow 0$ \\
    \For {\text{iteration} $=1, \dots, N$}{
        $B_{\text{t}} \sim X_{\text{t}}$, $B_{\text{nt}} \sim X_{\text{nt}}$ : $|B_{\text{t}}| = |B_{\text{nt}}| = \frac{b}{2}$ \\
        $\mathcal{L}_w = \alpha \mathcal{L}_{\text{t}} + \beta \mathcal{L}_{\text{nt}}$ \\
        $g_\delta \leftarrow \nabla_\delta \mathcal{L}_w$  \Comment*[r]{Calculate gradient} \ 
        $\delta \leftarrow$ \FAdam{$g_\delta$}  \Comment*[r]{Update perturbation} \
        $\delta \leftarrow \frac{\delta}{||\delta||_p}$ \Comment*[r]{Project to $l_p$ ball}
        }
\caption{Class Discriminative Universal Perturbation Generation}
\label{alg:cd_uap}
\end{algorithm}

\begin{table}[t]
\centering
\caption{Experiments on CIFAR100 for the ablation study of the proposed algorithm framework. Targeted classes are from 0 to 4. ($Acc_{\text{t}} = 65.20$; $Acc_{\text{nt}} = 70.43$).}
\label{tab:cifar100_data_split}
\small
    \begin{tabular}{lcccc}
    \toprule
    $X$                   & $\alpha$ & $\beta$ & $AAD_{\text{t}}$ & $AAD_{\text{nt}}$ \\ 
    \midrule
    with half-half        & $1$  & $1$            & $48.80$   & $17.31$\\
    without half-half     & $1$  & $1$            & $8.00$    & $1.79$\\
    without half-half     & $19$ & $1$            & $48.20$  & $19.43$ \\
    without half-half     & $1$  & $\frac{1}{19}$ & $47.00$  & $19.57$\\
    only targeted classes & $1$  & $0$            & $63.20$  & $53.67$\\
    \bottomrule
    \end{tabular}
\end{table}

Our goal is to design an algorithm achieving efficient generation of a class discriminative universal perturbation. 
Referring to the algorithm to generate universal adversarial perturbations introduced in~\cite{moosavi2017universal}, two limitations can be identified with regard to our specific problem: (1) the algorithm speed and (2) its non-discriminative nature. The algorithm seeks the perturbation $\delta$, with the minimal norm that allows to fool the network for a single data point $x$. This process is repeated over the training dataset until a certain fooling ratio is achieved while the perturbations are accumulated. However, despite its effectiveness, this approach does not leverage the power of parallel computing devices, such as GPUs, since in every iteration only a single image is processed. 
In our case, we speed up the perturbation crafting process with mini-batch training~\cite{goodfellow2016deep}.

To ensure that the generated universal perturbation is class-discriminative, a straightforward solution is to include only the samples belonging to the targeted class in the training process. One might expect the generated perturbation to fool the classifier only for samples from the targeted classes. However, a perturbation crafted only on the targeted classes deteriorates classification accuracy significantly for the non-targeted classes as well. 
The theoretical reason behind this observation is beyond the scope of this work, however, one clear take-away is that we need to exploit images from both targeted classes and non-targeted classes to achieve the desired goal of class discrimination. 

Our algorithm framework, explicitly assigning separate loss functions to the targeted and the non-targeted classes, is shown in Algorithm~\ref{alg:cd_uap}. As in most existing attack methods, gradients for perturbation updates are calculated with the standard backward propagation process using an optimizer. We empirically found that the widely used ADAM~\cite{kingma2014adam,reddy2018ask} optimizer converges faster than standard SGD. 
\begin{table}[t]
\centering
\caption{Experiments on CIFAR10 and CIFAR100 for various choices of loss functions. The two accuracies in each entry show the $AAD_{\text{t}}$ and $AAD_{\text{nt}}$. The initial performances for CIFAR10 on ResNet20 are $Acc_{\text{t}} = 90.86$, $Acc_{\text{nt}} = 92.44$; and for CIFAR100 on ResNet56 are $Acc_{t} = 65.20$; $Acc_{\text{nt}} = 70.43$, for targeted classes $0$ to $4$}
    \small
    \setlength\tabcolsep{1.3pt}
    \begin{tabular}{c|c|cc|cc|cc|cc}
    \toprule
    \multirow{2}{*}{CIFAR} & \multirow{2}{*}{$\mathcal{L}_{t}$} & \multicolumn{8}{c}{$\mathcal{L}_{\text{nt}}$} \\ 
    &    & \multicolumn{2}{c}{-} & \multicolumn{2}{c}{$\mathcal{L^{CE}}$} & \multicolumn{2}{c}{$\mathcal{L^{L}}$} & \multicolumn{2}{c}{$\mathcal{L^{BL}}$} \\
    \midrule
    \multirow{3}{*}{10} &
    $\mathcal{L}^{CE}$  & $87.04$  & $69.40$ & $84.36$  & $48.48$ & $84.68$  & $52.52$ & $83.86$  & $45.06$  \\
    &$\mathcal{L^{L}}$   & $84.64$  & $54.04$ & $85.74$  & $24.14$ & $81.04$  & $25.08$ & $84.94$  & $23.78$  \\
    &$\mathcal{L^{BL}}$  & $88.58$  & $61.54$ & $81.86$  & $10.18$ & $64.66$  & $7.92$  & $81.52$  & $11.60$  \\ 
    \midrule
    \multirow{3}{*}{100} & 
    $\mathcal{L}^{CE}$  & $61.20$  & $60.23$ & $60.40$  & $50.50$ & $1.60$   & $1.15$  & $62.40$  & $46.35$ \\
    &$\mathcal{L^{L}}$   & $63.80$  & $55.00$ & $63.40$  & $47.12$ & $62.40$  & $46.02$ & $62.80$  & $47.97$ \\
    &$\mathcal{L^{BL}}$  & $63.20$  & $53.67$ & $48.80$  & $17.31$ & $23.40$  & $7.16$  & $48.00$  & $17.79$ \\ 
    \bottomrule
    \end{tabular}
\label{tab:cifar_loss_choices}
\end{table}

One main characteristic of Algorithm~\ref{alg:cd_uap} is the 'half-half' batch data distribution strategy: half of the batch samples are randomly chosen from the targeted classes, while the other half is sampled from the non-targeted classes. This strategy is adopted to avoid imbalance in the batch data distribution. 
To illustrate this, we perform different experiments with and without the 'half-half' batch sampling strategy and different loss weighting parameters $\alpha$ and $\beta$ to compensate for data imbalance. The results are reported in Table~\ref{tab:cifar100_data_split}, with the best loss function configuration found, as discussed in the next subsection. 
One na\"{i}ve batch sampling approach could be randomly selecting the samples from all classes without distinguishing targeted classes and non-targeted classes. Since the ratio of the targeted classes to non-targeted classes in the training dataset is 5/95 (i.e.\ 1/19), much more samples would be chosen from the non-targeted classes, thereby dominating the targeted classes. In this case, we observe that both $AAD_{\text{t}}$ and $AAD_{\text{nt}}$ are very small. Note that changing $\alpha$ or $\beta$ proportional to the data ratio of targeted and non-targeted samples can mitigate this dominance of the non-targeted classes, however, the performance is still slightly worse than our proposed ``half-half" strategy. Another merit of the ``half-half" strategy is to facilitate the choice of weight parameters in Eq.~\ref{eq:loss}. Moreover, using only the samples of targeted classes for training also significantly deteriorates the model performance on the non-targeted classes. 

Another core part of the algorithm design is the exploration of different loss function configurations. 

\subsection{Loss Function Design}
\label{subsec:loss_fn}
\begin{table}[t]
\caption{Experiments on CIFAR10 with different groups of targeted classes using VGG16 and ResNet20}
\centering
\small
\setlength\tabcolsep{5.5pt}
    \begin{tabular}{ccccccc}
    \toprule
    $\hat{F}$      & $S$       & $Acc_{\text{t}}$ &  $AAD_{\text{t}}$  & $Acc_{\text{nt}}$ & $AAD_{\text{nt}}$ & $\Delta_{AAD}$ \\ 
    \midrule
    \parbox[t]{3mm}{\multirow{6}{*}{\rotatebox[origin=c]{90}{VGG16}}}
                          & $[1:5:2]$ & $90.57$ & $78.63$ & $94.23$ & $14.74$ & $\textbf{63.89}$ \\
                          & $[2:6:2]$ & $92.87$ & $68.87$ & $93.24$ & $21.79$ & $\textbf{47.08}$ \\
                          & $[0:4:1]$ & $92.40$ & $75.00$ & $93.86$ & $7.36$  & $\textbf{67.64}$ \\
                          & $[5:9:1]$ & $93.86$ & $75.00$ & $92.40$ & $17.56$ & $\textbf{57.44}$ \\
                          & $[0:6:1]$ & $92.10$ & $69.24$ & $95.53$ & $3.13$  & $\textbf{66.11}$ \\
                          & $[3:9:1]$ & $92.80$ & $78.60$ & $93.90$ & $8.70$  & $\textbf{69.90}$ \\ 
    \midrule
    \parbox[t]{3mm}{\multirow{6}{*}{\rotatebox[origin=c]{90}{ResNet20}}} 
                          & $[1:5:2]$ & $88.80$ & $82.27$ & $92.87$  & $15.93$ & $\textbf{66.34}$ \\
                          & $[2:6:2]$ & $92.30$ & $79.33$ & $91.37$  & $21.40$ & $\textbf{57.93}$ \\
                          & $[0:4:1]$ & $90.86$ & $81.46$ & $92.44$  & $10.24$ & $\textbf{71.22}$ \\
                          & $[5:9:1]$ & $92.44$ & $80.16$ & $90.86$  & $17.36$ & $\textbf{62.80}$ \\
                          & $[0:6:1]$ & $90.81$ & $81.33$ & $93.60$  & $4.67$ & $\textbf{76.66}$ \\
                          & $[3:9:1]$ & $91.21$ & $80.99$ & $92.67$  & $7.90$ & $\textbf{73.09}$ \\ 
    \bottomrule
    \end{tabular}
\label{tab:cifar10_results}
\end{table}

The loss function design is guided by the following intuitive principles. 
(1) For samples from the targeted classes, the loss function should guide the perturbation to fool the network. This can be realized through decreasing the logit value for the corresponding predicted class and optionally increasing the logit values of the remaining classes.
(2) For samples from the non-targeted classes, the loss function should guide the perturbation such that the logit of the predicted class remains the highest logit. 
(3) The objectives of (1) and (2) stay in conflict with each other, therefore, the loss functions for both parts need to be designed to have moderate influence on the gradient update, to avoid dominance of one part over the other. Taking objectives (1) and (2) into account, we deem it appropriate to separate the loss function into two parts for the samples from the targeted classes and those from the non-targeted classes, shown as
\begin{equation}
    \mathcal{L} = 
    \begin{cases}
         \mathcal{L}_{\text{t}} (x_{\text{t}}) & \text{ for } x_{\text{t}} \sim X_{\text{t}}\\
         \mathcal{L}_{\text{nt}} (x_{\text{nt}}) & \text{ for } x_{\text{nt}} \sim X_{\text{nt}.}\\
    \end{cases}
\end{equation}
Thus, the weighted loss $L_w$ can be expressed as
\begin{equation}
    \label{eq:loss}
    \mathcal{L}_w = \alpha \mathcal{L}_{\text{t}} + \beta \mathcal{L}_{\text{nt}}.
\end{equation}
As discussed in Table~\ref{tab:cifar100_data_split}, using the 'half-half'-strategy and setting $\alpha$ and $\beta$ to $1$ are appropriate design choices.
In practice, the attack hyper-parameters can be tailored to specific needs.
For example, an attacker can increase the parameter $\beta$ to get a more stealthy attack, consequently the attack success rate for the targeted classes will decrease. 
In the following section, we elaborate different variants of the loss functions for $\mathcal{L}_{\text{t}}$ and $\mathcal{L}_{\text{nt}}$. For simplicity, we only indicate the loss part for $\mathcal{L}_{\text{t}}$, since in the most na\"{i}ve form, $\mathcal{L}_{\text{nt}}$ can be achieved through a simple sign change $\mathcal{L}_{\text{nt}} = -\mathcal{L}_{\text{t}}$. However, we empirically found that this does not always provide the optimal solution, compared to the combination of different loss variants for $\mathcal{L}_{\text{t}}$ and $\mathcal{L}_{\text{nt}}$.

The cross-entropy loss, here indicated as $\mathcal{H}(\cdot)$, is a widely used loss function for training neural networks, and can be adapted for training a CD-UAP as follows:
\begin{equation}
\label{eq:ce_ce_loss}
    \mathcal{L}_{\text{t}}^{CE} =  - \mathcal{H}(\hat{F}(x + \delta), \hat{F}(x)).
\end{equation}
However, this formulation is prone to suffer from the property of the cross-entropy function, which takes logits of all classes into account. 
Reflecting on principle (3), we propose another loss function that directly operates on the logit values of the corresponding class in a more explicit way:
\begin{equation}
\label{eq:logit_loss}
    \mathcal{L}_{t}^{L} = \hat{L}_c(x + \delta),
\end{equation}
where $\hat{L}_c(\cdot)$ indicates the logit value of the predicted class $c = \argmax \hat{F}(x)$. 

Eq.~\ref{eq:logit_loss} has the drawback that optimization for the corresponding logits is unbounded. For a well trained network, we speculate that decreasing the logit of the corresponding class through a perturbation should be easier than increasing it. Thus, $\mathcal{L}_{t}$ is expected to dominate over $\mathcal{L}_{\text{nt}}$, which is supported by our experimental results (see Table~\ref{tab:cifar_loss_choices}). This problem can be mitigated by modifying the above loss function through a bounded logit expression: 
\begin{equation}
\label{eq:cwlogit_ce_loss}
    \mathcal{L}_{\text{t}}^{BL} = (\hat{L}_c(x + \delta) - \underset{i \neq c}\max \hat{L}_i(x + \delta))^+,
\end{equation}
where $(s)^+ = \max(s,0)$.

\begin{table}[t]
\centering
\caption{Experiments on CIFAR100 with different groups of targeted classes using VGG19 and ResNet56}
\label{tab:cifar100_results}
\small
\setlength\tabcolsep{4.5pt}
    \begin{tabular}{ccccccc}
    \toprule
                $\hat{F}$      & $S$ & $Acc_{\text{t}}$ & $AAD_{\text{t}}$ & $Acc_{\text{nt}}$ & $AAD_{\text{nt}}$ & $\Delta_{AAD}$ \\
    
    \midrule
    \parbox[t]{3mm}{\multirow{6}{*}{\rotatebox[origin=c]{90}{VGG 19}}}
                              & $[0:4:1]$    & $68.00$ & $54.50$ & $70.35$ & $17.04$ & $\textbf{37.46}$ \\
                              & $[10:50:10]$ & $63.00$ & $44.00$ & $70.61$ & $17.45$ & $\textbf{26.55}$ \\
                              & $[0:9:1]$    & $73.70$ & $42.70$ & $69.84$ & $22.57$ & $\textbf{20.13}$ \\
                              & $[0:90:10]$  & $69.10$ & $43.70$ & $70.36$ & $21.63$ & $\textbf{22.07}$ \\
                              & $[0:20:1]$   & $70.60$ & $36.70$ & $70.14$ & $23.27$ & $\textbf{13.43}$ \\
                              & $[0:95:5]$   & $68.20$ & $38.65$ & $70.34$ & $23.30$ & $\textbf{15.35}$ \\ 
    \midrule
    \parbox[t]{3mm}{\multirow{6}{*}{\rotatebox[origin=c]{90}{ResNet 56}}}
                              & $[0:4:1]$    & $65.20$ & $49.00$ & $70.43$ & $16.54$ & $\textbf{32.46}$ \\
                              & $[10:50:10]$ & $61.80$ & $41.80$ & $70.61$ & $16.54$ & $\textbf{25.26}$ \\
                              & $[0:9:1]$    & $71.30$ & $39.70$ & $70.04$ & $21.96$ & $\textbf{17.74}$ \\
                              & $[0:90:10]$  & $68.10$ & $41.10$ & $70.40$ & $19.95$ & $\textbf{21.15}$ \\
                              & $[0:20:1]$   & $68.85$ & $34.90$ & $70.50$ & $22.36$ & $\textbf{12.54}$ \\
                              & $[0:95:5]$   & $68.20$ & $35.25$ & $70.66$ & $22.05$ & $\textbf{13.20}$ \\ 
    \bottomrule
    \end{tabular}
\end{table}

\begin{table*}[t]
\centering
\caption{Experiments on CIFAR100 targeting superclasses using VGG19 and ResNet56}
\label{tab:cifar100_superclasses_results}
\small
\begin{tabular}{l|ccccc|ccccc}
\toprule
\multirow{2}{*}{Super Class} & \multicolumn{5}{c}{VGG19}   & \multicolumn{5}{c}{ResNet56}   \\
& $Acc_{\text{t}}$ & $AAD_{\text{t}}$          & $Acc_{\text{nt}}$ & $AAD_{\text{nt}}$    & $\Delta_{AAD}$ & $Acc_{\text{t}}$ & $AAD_{\text{t}}$  & $Acc_{\text{nt}}$ & $AAD_{\text{nt}}$   & $\Delta_{AAD}$ \\
\midrule
aquatic mammals                    & $56.20$ & $44.20$ & $70.97$ & ${14.05}$ & $\textbf{30.15}$ & $58.20$ & ${43.60}$ & $70.80$ & ${14.87}$ & $\textbf{28.73}$ \\
fish                               & $67.00$ & ${45.60}$ & $70.40$ & ${18.25}$ & $\textbf{27.35}$ & $67.80$ & ${49.00}$ & $70.30$ & ${18.45}$ & $\textbf{30.55}$ \\
flowers                            & $76.40$ & ${33.80}$ & $69.91$ & ${20.57}$ & $\textbf{13.23}$ & $75.40$ & ${30.80}$ & $69.90$ & ${24.18}$ &  $\textbf{6.62}$ \\
food containers                    & $69.60$ & ${52.40}$ & $70.26$ & ${16.28}$ & $\textbf{36.12}$ & $71.40$ & ${54.40}$ & $70.11$ & ${16.44}$ & $\textbf{37.96}$ \\
fruit and vegetables               & $79.40$ & ${55.60}$ & $69.75$ & ${18.04}$ & $\textbf{37.56}$ & $76.60$ & ${52.00}$ & $69.83$ & ${17.65}$ & $\textbf{34.35}$ \\
household electrical devices       & $70.00$ & ${49.20}$ & $70.24$ & ${15.95}$ & $\textbf{33.25}$ & $75.00$ & ${55.60}$ & $69.92$ & ${16.52}$ & $\textbf{39.08}$ \\
household furniture                & $77.40$ & ${63.80}$ & $69.85$ & ${15.76}$ & $\textbf{48.04}$ & $76.00$ & ${60.00}$ & $69.86$ & ${14.90}$ & $\textbf{45.10}$ \\
insects                            & $70.20$ & ${38.60}$ & $70.23$ & ${16.99}$ & $\textbf{21.61}$ & $71.60$ & ${45.20}$ & $70.09$ & ${21.24}$ & $\textbf{23.96}$ \\
large carnivores                   & $70.20$ & ${53.60}$ & $70.23$ & ${15.50}$ & $\textbf{38.10}$ & $70.40$ & ${60.20}$ & $70.16$ & ${15.44}$ & $\textbf{44.76}$ \\
large man-made outdoor objects     & $82.60$ & ${66.60}$ & $69.58$ & ${15.56}$ & $\textbf{51.04}$ & $81.20$ & ${66.20}$ & $69.59$ & ${15.49}$ & $\textbf{50.71}$ \\
large natural outdoor scenes       & $79.00$ & ${70.80}$ & $69.77$ & ${13.12}$ & $\textbf{57.68}$ & $78.40$ & ${67.60}$ & $69.74$ & ${12.13}$ & $\textbf{55.47}$ \\
large omnivores and herbivores     & $69.80$ & ${51.60}$ & $70.25$ & ${18.33}$ & $\textbf{33.27}$ & $71.40$ & ${56.00}$ & $70.10$ & ${17.15}$ & $\textbf{38.85}$ \\
medium-sized mammals               & $72.80$ & ${49.40}$ & $70.09$ & ${17.27}$ & $\textbf{32.13}$ & $71.80$ & ${54.20}$ & $70.08$ & ${18.83}$ & $\textbf{35.37}$ \\
non-insect invertebrates           & $66.40$ & ${40.60}$ & $70.43$ & ${18.09}$ & $\textbf{22.51}$ & $66.00$ & ${44.60}$ & $70.39$ & ${18.38}$ & $\textbf{26.22}$ \\
people                             & $52.20$ & ${31.60}$ & $71.18$ & ${14.67}$ & $\textbf{16.93}$ & $48.40$ & ${32.60}$ & $71.32$ & ${14.85}$ & $\textbf{17.75}$ \\
reptiles                           & $59.20$ & ${43.80}$ & $70.81$ & ${16.05}$ & $\textbf{27.76}$ & $59.20$ & ${42.80}$ & $70.74$ & ${17.24}$ & $\textbf{25.56}$ \\
small mammals                      & $56.00$ & ${47.60}$ & $70.98$ & ${15.14}$ & $\textbf{32.46}$ & $56.40$ & ${47.20}$ & $70.90$ & ${13.34}$ & $\textbf{33.86}$ \\
trees                              & $66.80$ & ${49.60}$ & $70.41$ & ${15.58}$ & $\textbf{34.02}$ & $66.60$ & ${54.00}$ & $70.36$ & ${17.05}$ & $\textbf{36.95}$ \\
vehicles 1                         & $82.20$ & ${44.00}$ & $69.60$ & ${17.99}$ & $\textbf{26.01}$ & $80.60$ & ${51.60}$ & $69.62$ & ${20.44}$ & $\textbf{31.16}$ \\
vehicles 2                         & $81.20$ & ${53.40}$ & $69.65$ & ${21.33}$ & $\textbf{32.07}$ & $81.00$ & ${62.20}$ & $69.60$ & ${21.92}$ & $\textbf{40.28}$ \\ 
\bottomrule
\end{tabular}
\end{table*}

\begin{table*}[t]
\centering
\caption{Experiments on CIFAR100 targeting 2 super classes simultaneously using VGG19 and ResNet56}
\label{tab:cifar100_2superclasses_results}
\small
\setlength\tabcolsep{4.0pt}
\begin{tabular}{l|ccccc|ccccccc}
\toprule
\multirow{2}{*}{Super Class} & \multicolumn{5}{c}{VGG19}   & \multicolumn{5}{c}{ResNet56} \\
& $Acc_{\text{t}}$ & $AAD_{\text{t}}$ & $Acc_{\text{nt}}$ & $AAD_{\text{nt}}$ & $\Delta_{AAD}$ & $Acc_{\text{t}}$ & $AAD_{\text{t}}$ & $Acc_{\text{nt}}$ & $AAD_{\text{nt}}$ & $\Delta_{AAD}$ \\
\midrule
aquatic mammals + fish                  & $61.60$ & $37.60$ & $71.19$ & $16.86$ & $\textbf{20.74}$ & $63.00$ & $39.40$ & $70.97$ & $18.26$ & $\textbf{21.14}$ \\
flowers + food containers               & $73.00$ & $29.70$ & $69.92$ & $16.77$ & $\textbf{12.93}$ & $73.40$ & $32.90$ & $69.81$ & $21.12$ & $\textbf{11.78}$ \\
fruit/vegetables + electronics          & $74.70$ & $43.80$ & $69.73$ & $17.69$ & $\textbf{26.11}$ & $75.80$ & $46.50$ & $69.54$ & $18.78$ & $\textbf{27.72}$ \\
household furniture + insects           & $73.80$ & ${41.30}$ & $69.83$ & ${17.33}$ & $\textbf{23.97}$ & $73.80$ & ${46.60}$ & $69.77$ & ${18.38}$ & $\textbf{28.22}$ \\
large carnivores + outdoors objects     & $76.40$ & ${51.80}$ & $69.54$ & ${16.62}$ & $\textbf{35.18}$ & $75.80$ & ${59.00}$ & $69.54$ & ${18.64}$ & $\textbf{40.36}$ \\
natural outdoors + omnivores/herbivores & $74.40$ & ${53.80}$ & $69.77$ & ${17.23}$ & $\textbf{36.57}$ & $74.90$ & ${56.30}$ & $69.64$ & ${17.02}$ & $\textbf{39.28}$ \\
medium-mammals + non-ins. invertebrates & $69.60$ & ${35.90}$ & $70.30$ & ${19.26}$ & $\textbf{16.64}$ & $68.90$ & ${44.20}$ & $70.31$ & ${21.38}$ & $\textbf{22.82}$ \\
people + reptiles                       & $55.70$ & ${30.40}$ & $71.84$ & ${16.79}$ & $\textbf{13.61}$ & $53.80$ & ${32.10}$ & $71.99$ & ${17.30}$ & $\textbf{14.80}$ \\
small mammals + trees                   & $61.40$ & ${42.40}$ & $71.21$ & ${16.48}$ & $\textbf{25.92}$ & $61.50$ & ${42.70}$ & $71.13$ & ${17.78}$ & $\textbf{24.92}$ \\
vehicles 1 \& 2                         & $81.70$ & ${46.00}$ & $68.96$ & ${19.78}$ & $\textbf{26.22}$ & $80.80$ & ${54.70}$ & $68.99$ & ${21.43}$ & $\textbf{33.27}$ \\ 
\bottomrule
\end{tabular}
\end{table*}

We explore the effect of different loss functions on CIFAR100 and report the results in Table~\ref{tab:cifar_loss_choices}. 
Three major observations can be made from the results in Table~\ref{tab:cifar_loss_choices}. First, when the loss function is applied only for the targeted classes, the crafted perturbation deteriorates network performance on both the targeted and non-targeted classes. More specifically, the $AAD_{\text{nt}}$ is only slightly lower than $AAD_{\text{t}}$, which shows that na\"{i}vely crafting the perturbation on images of the targeted classes cannot generate a class-discriminative UAP. Second, for the targeted classes, the effect of both $\mathcal{L^{CE}}$ and $\mathcal{L^{L}}$ is relatively dominant, and thus detrimental to model performance for samples of non-targeted classes. Third, the effect of $\mathcal{L^{BL}}$ is more moderate than $\mathcal{L^{L}}$ for both samples from targeted classes and non-targeted classes, which makes $\mathcal{L^{BL}}$ a more appropriate loss function, especially for the targeted classes.

Based on the above observations, we choose $\mathcal{L^{BL}}$ as the loss function $\mathcal{L}_{t}$. 
For samples from the non-targeted classes, we observe that $\mathcal{L^{CE}}$ outperforms $\mathcal{L^{BL}}$ with a small margin (i.e.\ yields a slightly higher $AAD_{\text{t}}$ and lower $AAD_{\text{nt}}$). The same phenomena can be observed for experiments of CIFAR10. Thus, we choose $\mathcal{L^{CE}}$ as the loss function $\mathcal{L}_{nt}$.

To sum up, we first give a definition of the task of CD-UAP and propose an algorithm framework catering for the practical needs of high efficiency and class-discrimination. We then design and compare different loss function configurations. 

\section{Experimental Results and Analysis}

Before presenting the results of our experiments, we briefly discuss our experimental setup.
The CD-UAP generated by the best-performing loss configuration found above is then extensively evaluated on three datasets for various network architectures.

\subsection{Implementation Details}
For CIFAR and ImageNet datasets, we deploy the $l_\infty$-norm on $\delta$ with $\epsilon=10$ and $\epsilon=15$, respectively, for natural images in the range of $[0,255]$. As discussed earlier, we use the ADAM optimizer for all experiments, setting the batch size to $128$ for CIFAR10 and CIFAR100~\cite{krizhevsky2009learning} experiments, and $32$ for experiments on ImageNet~\cite{deng2009imagenet}. In all our experiments, we train the CD-UAP on the training dataset. Specifically, we only use the initially correctly classified samples in the training dataset. The generated CD-UAP is evaluated on the test dataset. All experiments are conducted with the PyTorch framework.

\subsection{Experimental Results}
\subsubsection{CIFAR10} 
The results for CD-UAPs on CIFAR10 with VGG16~\cite{simonyan2014very} and ResNet20~\cite{he2016deep} are available in Table~\ref{tab:cifar10_results}. The targeted classes are listed under $S$ in the second column, in the format [\textit{first class index : last class index : step size}]. For example, $[1:5:2]$ indicates that classes $1$, $3$ and $5$ are selected as the targeted classes.
We observe that for the same trained model, there is a visible variation when different groups of classes are chosen. Nonetheless, for both VGG16 and ResNet20, the significant gap $\Delta_{AAD}$ between $AAD_{t}$ and $AAD_{nt}$ shows the effectiveness of the proposed approach. 

\subsubsection{CIFAR100}
\begin{figure}[t]
\includegraphics[width=0.45\textwidth]{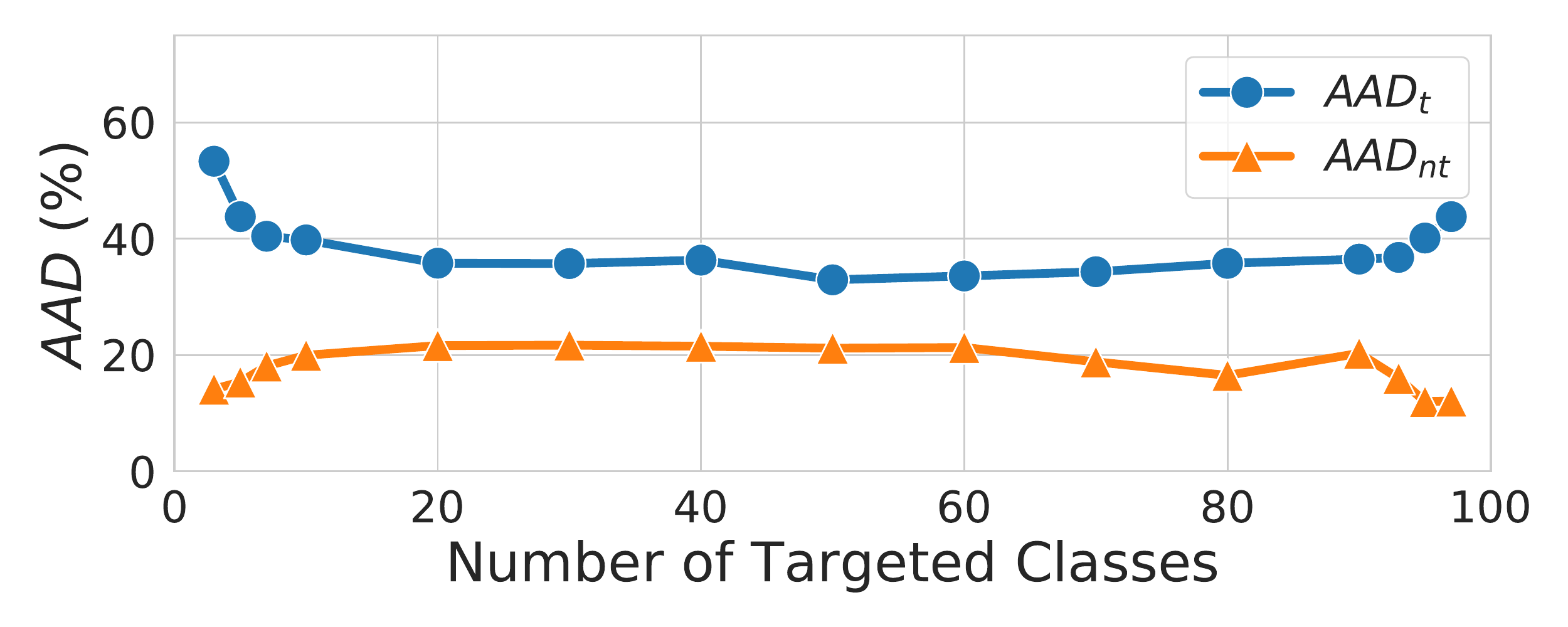}
\caption{$AAD_{\text{t}}$ and $AAD_{\text{nt}}$ over the number of targeted classes on CIFAR100}
\label{fig:acc_over_number_of_classes}
\end{figure}
Furthermore, we evaluate CD-UAP on CIFAR100 and report the results in Table~\ref{tab:cifar100_results}. We observe similar trends on CIFAR100 as for CIFAR10. The overall performance is relatively lower than that for CIFAR10 due to the increasing complexity of the task. Specifically, we observe that the performance decreases with the increase of the number of targeted classes. In Figure~\ref{fig:acc_over_number_of_classes} we further investigate the influence of the number of targeted classes on the CD-UAP performance.
The results show that CD-UAP performs best with either a low (up to $10$) or a high (above $90$) number of targeted classes with a relatively lower performance in between.

CIFAR100 has semantically similar classes which can be grouped into 20 super classes, each consisting of 5 sub-classes. For example, the super class fish comprises of aquarium fish, flatfish, ray, shark and trout. We argue that it is practically meaningful to attack super classes as a group instead of targeting random classes. The results for targeting one super class on CIFAR100 are shown in Table~\ref{tab:cifar100_superclasses_results}. We observe a reasonably large gap $\Delta_{AAD}$ for all super classes, with visible variations for different super classes. 
Attacking two super classes simultaneously is explored in Table~\ref{tab:cifar100_2superclasses_results}. Attacking multiple super classes performs inferior to one super class, indicating that it is harder to craft a CD-UAP with an increasing variation among the targeted classes. 

\begin{table}[t]
\centering
\caption{Experiments on ImageNet targeting 1 super class, $\epsilon=15$.}
\label{tab:imagenet_results}
\small
\setlength\tabcolsep{1.7pt}
\scalebox{1.0}{
    \begin{tabular}{lcc ccc c}
    \toprule
           & Super Classes       &      $Acc_{\text{t}}$     & $AAD_{\text{t}}$ &   $Acc_{\text{nt}}$        & $AAD_{\text{nt}}$ & $\Delta_{AAD}$ \\ 
    \midrule
        \parbox[t]{3mm}{\multirow{5}{*}{\rotatebox[origin=c]{90}{VGG16}}}
    & Frogs                       & $70.0$ & $46.0$ & $71.6$ & $19.5$ & $\textbf{26.5}$ \\
    & Sharks                      & $80.0$ & $53.3$ & $71.6$ & $16.8$ & $\textbf{36.5}$ \\
    & Aircrafts                   & $78.0$ & $69.3$ & $71.6$ & $17.7$ & $\textbf{51.6}$ \\ 
    & Racket Radiator Radio       & $68.6$ & $42.7$ & $71.6$ & $18.5$ & $\textbf{24.2}$ \\
    & Space objects                 & $56.7$ & $20.0$ & $71.6$ & $18.5$ & $\textbf{1.5}$ \\
    \midrule
        \parbox[t]{3mm}{\multirow{5}{*}{\rotatebox[origin=c]{90}{VGG19}}} 
    & Frogs                       & $70.0$ & $42.0$ & $72.4$ & $18.0$ & $\textbf{24.0}$ \\
    & Sharks                      & $81.3$ & $55.3$ & $72.3$ & $16.1$ & $\textbf{39.2}$ \\
    & Aircrafts                   & $81.3$ & $72.0$ & $72.4$ & $17.2$ & $\textbf{54.8}$ \\
    & Racket Radiator Radio       & $69.3$ & $37.3$ & $72.4$ & $17.3$ & $\textbf{20.0}$ \\
    & Space objects                 & $59.3$ & $26.6$ & $72.4$ & $20.0$ & $\textbf{6.6}$  \\
    \midrule
        \parbox[t]{3mm}{\multirow{5}{*}{\rotatebox[origin=c]{90}{ResNet50}}}
    & Frogs                       & $75.3$ & $52.0$ & $76.1$ & $18.5$ & $\textbf{33.5}$ \\
    & Sharks                      & $83.3$ & $66.7$ & $76.1$ & $16.1$ & $\textbf{50.6}$ \\
    & Aircrafts                   & $84.0$ & $65.3$ & $76.1$ & $16.8$ & $\textbf{48.5}$ \\ 
    & Racket Radiator Radio       & $75.3$ & $46.7$ & $76.1$ & $17.1$ & $\textbf{29.6}$ \\
    & Space objects                 & $59.3$ & $43.3$ & $76.2$ & $20.4$ & $\textbf{22.9}$ \\
    \midrule
        \parbox[t]{3mm}{\multirow{5}{*}{\rotatebox[origin=c]{90}{ResNet152}}}
    & Frogs                       & $74.0$ & $48.0$ & $78.3$ & $17.2$ & $\textbf{30.8}$ \\
    & Sharks                      & $80.0$ & $61.3$ & $78.3$ & $12.9$ & $\textbf{48.4}$ \\
    & Aircrafts                   & $86.0$ & $78.7$ & $78.3$ & $16.1$ & $\textbf{62.6}$ \\
    & Racket Radiator Radio       & $72.7$ & $44.0$ & $78.3$ & $14.4$ & $\textbf{29.6}$ \\
    & Space objects                 & $64.7$ & $36.7$ & $78.4$ & $16.7$ & $\textbf{20.0}$ \\ 
    \bottomrule
    \end{tabular}
}
\end{table}

\begin{table}[t]
\centering
\caption{AC-UAP task performance compared with UAP~\cite{moosavi2017universal} and GAP~\cite{poursaeed2018generative}.}
\small
\setlength\tabcolsep{1.7pt}
\scalebox{1.0}{
\label{tab:uap}
    \begin{tabular}{lcccc}
    \toprule
        & VGG16 & VGG19 & ResNet152 & Inception-V3 \\
        UAP & $78.3$ & $77.8$ &   $84.0$   & $-$ \\
        GAP & $83.7$ & $80.1$ &  -         & $82.7$ \\
    \midrule
        CD-UAP ($\mathcal{L}^{\text{CE}}$)   & $93.1$ & $93.5$ & $86.8$ & $83.1$ \\
        CD-UAP ($\mathcal{L}^{\text{BL}}$)   & $\textbf{93.7}$ & $\textbf{94.2}$ & $\textbf{90.2}$ & $\textbf{85.9}$ \\
    \bottomrule
    \end{tabular}
    }
\end{table}

\begin{table}[t]
\centering
\caption{Transferability Experiments. $^*$ indicates the white-box CD-UAP.}
\label{tab:transferability}
\small
\setlength\tabcolsep{1.7pt}
\begin{tabular}{cclllll}
\toprule
$\hat{F}_s$      & $\hat{F}_t$ & $Acc_{\text{t}}$ &  $AAD_{\text{t}}$  & $Acc_{\text{nt}}$ & $AAD_{\text{nt}}$ & $\Delta_{AAD}$ \\ 
\midrule
\multirow{4}{*}{VGG16} 
                          & VGG16       & $78.00^*$ & $69.33^*$ & $71.57^*$  & $17.68^*$ & $\textbf{51.65}^*$\\
                          & VGG19       & $78.00$ & $40.00$ & $71.57$  & $14.08$ & $\textbf{25.92}$ \\
                          & ResNet50    & $78.00$ & $ 2.00$ & $71.57$  & $ 1.57$ & $\textbf{0.43}$\\ 
                          & ResNet152   & $78.00$ & $ 0.67$ & $71.57$  & $ 0.02$ & $\textbf{0.65}$\\ 
\midrule
\multirow{4}{*}{VGG19}
                          & VGG16       & $81.33$ & $47.33$ & $72.35$  & $16.41$ & $\textbf{30.92}$ \\
                          & VGG19       & ${81.33}^*$ & ${72.00}^*$ & ${72.35}^*$  & $17.19^*$ & $\textbf{54.81}^*$  \\
                          & ResNet50    & $81.33$ & $ 7.33$ & $72.35$  & $ 5.49$ &$\textbf{1.84}$\\
                          & ResNet152   & $81.33$ & $ 5.33$ & $72.35$  & $ 0.94$ &$\textbf{4.39}$\\ 
\midrule
\multirow{4}{*}{ResNet50} 
                          & VGG16       & $84.00$ & $47.33$ & $76.11$  & $22.08$ &$\textbf{25.25}$\\
                          & VGG19       & $84.00$ & $40.67$ & $76.11$  & $16.82$ &$\textbf{23.85}$\\ 
                          & ResNet50    & ${84.00}^*$ & ${65.33}^*$ & ${76.11}^*$  & ${16.82}^*$ &$\textbf{48.51}^*$\\ 
                          & ResNet152   & $84.00$ & $20.67$ & $76.11$  & $ 6.88$ &$\textbf{13.79}$\\ 
\midrule
\multirow{4}{*}{ResNet152} 
                          & VGG16       & $86.00$ & $54.67$ & $78.29$  & $25.41$ &$\textbf{29.26}$\\
                          & VGG19       & $86.00$ & $48.67$ & $78.29$  & $23.86$ &$\textbf{24.81}$\\
                          & ResNet50    & $86.00$ & $34.67$ & $78.29$  & $15.18$ &$\textbf{19.49}$\\ 
                          & ResNet152   & ${86.00}^*$ & ${78.67}^*$ & ${78.29}^*$  & ${16.08}^*$ &$\textbf{62.59}^*$\\ 
\bottomrule
\end{tabular}
\end{table}

\subsubsection{ImageNet}
The results for evaluating CD-UAP on ImageNet are available in Table~\ref{tab:imagenet_results}. 
For this experiment, we attack four state-of-the-art networks for five different super classes, each comprising of three sub-classes. All the results consistently show a higher $AAD_t$ than $AAD_{nt}$, resulting in a non-trivial gap ($\Delta_{AAD}$) between them. 

Data availability can be a concern in practice. We further explore the performance of CD-UAP under limited data-availability, using 100 images per class in the training dataset (less than 10\% of the whole dataset). For the superclass of Aircrafts on ResNet50, the CD-UAP can achieve an absolute accuracy drop of $52.00\%$ and $14.55\%$ for $AAD_{\text{t}}$ and $AAD_{\text{nt}}$, respectively. Even though less than 10\% of the whole training dataset are used, the CD-UAP still achieves reasonable performance. 

\subsubsection{Performance Comparison for AC-UAP}
AC-UAP attacks all classes and can be seen as a special case of the proposed CD-UAP when all classes are targeted. For this special case, we compare our proposed approach with the existing UAP methods: UAP~\cite{moosavi2017universal} and GAP~\cite{poursaeed2018generative} in Table~\ref{tab:uap}. We observe that our proposed approach (with the same constraint $\epsilon=10$ as UAP and GAP) outperforms the existing methods by a significant margin, achieving state-of-the-art performance for the task of AC-UAP. Note that our approach is much more efficient than UAP since we do not deploy the cumbersome DeepFool algorithm, and our approach does not require training of another network as GAP.

\subsubsection{Qualitative Results}
\begin{figure}[t]
\centering
\small
\includegraphics[width=\linewidth]{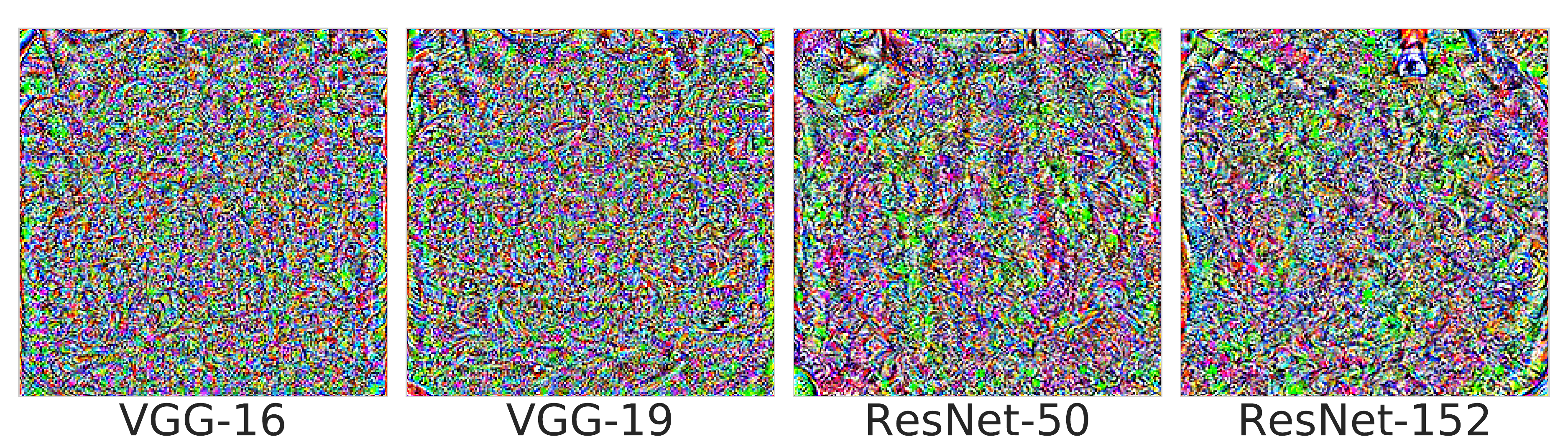}
\caption{CD-UAP generated for different networks}
\label{fig:pert}
\end{figure}

The generated CD-UAPs on ImageNet are amplified and visualized in Figure \ref{fig:pert}. Since the magnitude of the perturbation is relatively small, adding it to the images will not produce changes perceptible to a human observer. Thus, we only report the perturbations themselves. The generated perturbation patterns are observed to somehow link to the network type. For VGG networks~\cite{simonyan2014very}, the crafted perturbations look like random noise, while those for ResNet tend to demonstrate some pattern, which however is not interpretable by a human observer.

\subsubsection{CD-UAP Transferability}
We report the CD-UAP transferability between different networks in Table~\ref{tab:transferability} from which there are two major observations. First, for two networks from the same network family, the CD-UAP tends to transfer well among them. For example, the CD-UAPs crafted for VGG16 and VGG19 from the VGGNet-family transfer well to each other. Second, for networks from different network families, the transferability sometimes fails. For example, ResNet can transfer well to VGGNet, but not vice versa. The reason of this phenomenon is left for future work.

\section{Conclusion}
Identifying the limitation of the existing UAP methods, we proposed class discriminative universal adversarial perturbation (CD-UAP), that aims to attack only images of the targeted classes, while having minimal influence on other classes. To generate such perturbation, we proposed a simple yet effective algorithm framework, which separately deals with samples from targeted and non-targeted classes. Under the proposed framework, we design and compare different loss function configurations to search for the optimal combination for targeted and non-targeted classes. The effectiveness of our approach is demonstrated through extensive experimentation on the CIFAR10, CIFAR100 and ImageNet datasets. Moreover, we found that in-general the task complexity of CD-UAP increases with the number of targeted classes. For the task of AC-UAP, our proposed approach achieves state-of-the-art performance, outperforming the existing methods by a significant margin. We further provide additional experiments demonstrating the transferability between different networks. 

\bibliographystyle{aaai}
\bibliography{bibliography}

\end{document}